\documentclass[10pt,twocolumn,letterpaper]{article}

\usepackage{3dv}
\usepackage{times}
\usepackage{epsfig}
\usepackage{graphicx}
\usepackage{amsmath}
\usepackage{amssymb}
\usepackage[dvipsnames]{xcolor}
\usepackage{subcaption}
\usepackage{pdfpages}

\usepackage[pagebackref=true,breaklinks=true,colorlinks,bookmarks=false]{hyperref}

\usepackage[font=small,labelfont=bf,tableposition=top]{caption}

\threedvfinalcopy % *** Uncomment this line for the final submission

\def\threedvPaperID{194} % *** Enter the 3DV Paper ID here

\begin{document}

%%%%%%%%% TITLE
\title{LM-Reloc: Levenberg-Marquardt Based Direct Visual Relocalization}

\author{Lukas von Stumberg$^\text{1,2*}$ \quad Patrick Wenzel$^\text{1,2*}$ \quad Nan Yang$^\text{1,2}$ \quad Daniel Cremers$^\text{1,2}$ \\
$^\text{1}$ Technical University of Munich \quad $^\text{2}$ Artisense}

%% TEASER FIGURE
\twocolumn[{%val
	\renewcommand\twocolumn[1][]{#1}%
	\maketitle
	\begin{center}
		\vspace{-0.55cm}
		\captionsetup{type=figure}
    \includegraphics[width=1.0\linewidth]{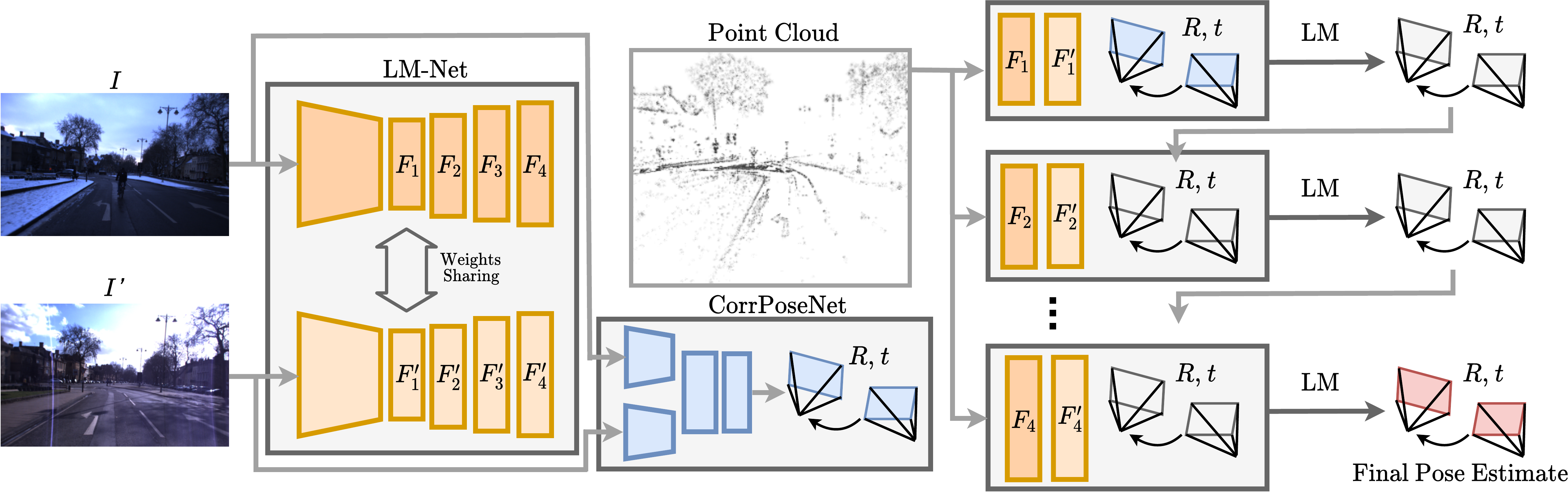}
		\captionof{figure}{
We propose LM-Reloc -- a novel approach for visual relocalization based on direct image alignment. It consists of two deep neural networks: LM-Net, an encoder-decoder network for learning dense visual descriptors and a CorrPoseNet to bootstrap the direct image alignment. The final 6DoF relative pose estimate between image $I$ and $I^{\prime}$ is obtained in a coarse-to-fine pyramid scheme leveraging the learned feature maps. The initialization for the direct image alignment is obtained by the CorrPoseNet.
    	}
    	\label{fig:teaser}
	\end{center}
}]
%%%

\renewcommand*{\thefootnote}{\fnsymbol{footnote}}
\footnotetext[1]{Equal contribution.}

%%%%%%%%% ABSTRACT
\begin{abstract}
\vspace{-0.41cm}
We present LM-Reloc -- a novel approach for visual relocalization based on direct image alignment. In contrast to prior works that tackle the problem with a feature-based formulation, the proposed method does not rely on feature matching and RANSAC. Hence, the method can utilize not only corners but any region of the image with gradients. In particular, we propose a loss formulation inspired by the classical Levenberg-Marquardt algorithm to train LM-Net. The learned features significantly improve the robustness of direct image alignment, especially for relocalization across different conditions. To further improve the robustness of LM-Net against large image baselines, we propose a pose estimation network, CorrPoseNet, which regresses the relative pose to bootstrap the direct image alignment. Evaluations on the CARLA and Oxford RobotCar relocalization tracking benchmark show that our approach delivers more accurate results than previous state-of-the-art methods while being comparable in terms of robustness.
\end{abstract}

%%%%%%%%% BODY TEXT
\section{Introduction}
Map-based relocalization, that is, to localize a camera within a pre-built reference map, is becoming more and more important for robotics~\cite{fabmap}, autonomous driving~\cite{ort2018autonomous,brubaker2015map} and AR/VR~\cite{8436423}. Sequential-based approaches, which leverage the temporal structure of the scene provide more stable pose estimations and also deliver the positions in global coordinates compared to single image-based localization methods. The map is usually generated by either using LiDAR or visual Simultaneous Localization and Mapping (vSLAM) solutions. In this paper, we consider \mbox{vSLAM} maps due to the lower-cost visual sensors and the richer semantic information from the images. Feature-based methods~\cite{klein2007parallel,davison2007monoslam,Mur-ArtalIEEETR2015,orbslam2} and direct methods~\cite{kerl2013dense,engel14eccv,dso,Alismail2016PhotometricBA} are two main lines of research for vSLAM.

Once a map is available, the problem of relocalizing within this map at any later point in time requires to deal with long-term changes in the environment. This makes a centimeter-accurate global localization challenging, especially in the presence of drastic lighting and appearance changes in the scene. For this task, feature-based methods are the most commonly used approaches to estimate the ego-pose and its orientation. This is mainly due to the advantage that features are more robust against changes in lighting/illumination in the scene.

However, feature-based methods can only utilize keypoints that have to be matched across the images before the pose estimation begins. Thus they ignore large parts of the available information. Direct methods, in contrast, can take advantage of all image regions with sufficient gradients and as a result, are known to be more accurate on visual odometry benchmarks~\cite{photopatch,dso,yang2018challenges}.

In this paper, we propose LM-Reloc, which applies direct techniques to the task of relocalization. LM-Reloc consists of LM-Net, CorrPoseNet, and a non-linear optimizer, which work seamlessly together to deliver reliable pose estimation without RANSAC and feature matching. In particular, we derive a loss formulation, which is specifically designed to work well with the Levenberg-Marquardt (LM) algorithm~\cite{levenberg1944method,marquardt1963algorithm}. We use a deep neural network, LM-Net, to train descriptors that are being fed to the direct image alignment algorithm. Using these features results in better robustness against bad initializations, large baselines, and against illumination changes.

While the robustness improvements gained with our loss formulation are sufficient in many cases, for very large baselines or strong rotations, some initialization can still be necessary. To this end, we propose a pose estimation network. Based on two images it directly regresses the 6DoF pose, which we utilize as initialization for LM-Net. The CorrPoseNet contains a correlation layer as proposed in~\cite{rocco2017convolutional}, which ensures that the network can handle large displacements. The proposed CorrPoseNet displays a lot of synergies with LM-Net. Despite being quite robust, the predictions of the CorrPoseNet are not very accurate. Thus it is best used in conjunction with our LM-Net, resulting in very robust and accurate pose estimates.

We evaluate our approach on the relocalization tracking benchmark from~\cite{gnnet}, which contains scenes simulated using CARLA~\cite{DosovitskiyCoRL2017}, as well as sequences from the Oxford RobotCar dataset~\cite{MaddernIJRR2017}. Our LM-Net shows superior accuracy especially in terms of rotation while being competitive in terms of robustness.

We summarize our main contributions:

\begin{itemize}
    \item LM-Reloc, a novel pipeline for visual relocalization based on direct image alignment, which consists of LM-Net, CorrPoseNet, and a non-linear optimizer.
    \item A novel loss formulation together with a point sampling strategy that is used to train LM-Net such that the resulting feature descriptors are optimally suited to work with the LM algorithm.
    \item Extensive evaluations on the CARLA and Oxford RobotCar relocalization tracking benchmark which show that the proposed approach achieves state-of-the-art relocalization accuracy without relying on feature matching or RANSAC.
\end{itemize}

\section{Related Work}
In this section, we review the main topics that are closely related to our work, including direct methods for visual localization and feature-based visual localization methods.

\noindent{\textbf{Direct methods for visual localization.}} In recent years, direct methods~\cite{kerl2013dense,engel14eccv,dso} for SLAM and visual odometry have seen a great progress. Unlike feature-based methods~\cite{klein2007parallel,davison2007monoslam,Mur-ArtalIEEETR2015,orbslam2} which firstly extracts keypoints as well as the corresponding descriptors, and then minimize the geometric errors, direct methods minimize the energy function based on the photometric constancy assumption without performing feature matching or RANSAC. By utilizing more points from the images, direct methods show higher accuracy than feature-based methods~\cite{yang2018challenges}. However, classical direct methods show lower robustness than feature-based methods when the photometric constancy assumption is violated due to, e.g., the lighting and weather changes which are typical for long-term localization~\cite{SattlerCVPR2018}. In~\cite{AlismailIEEERAL2017} and~\cite{pascoe2017nid}, the authors propose to use the handcrafted features to improve the robustness of direct methods against low light or global appearance changes. Some recent works~\cite{chang2017clkn,lv2019taking,gnnet} address the issue by using learned features from deep neural networks~\cite{lecun2015deeplearning}. In~\cite{chang2017clkn} they train deep features using a Hinge-Loss based on the Lucas-Kanade method, however, in contrast to us, they estimate the optical flow instead of applying the features to the task of relocalization. The most related work to ours is GN-Net~\cite{gnnet} which proposes a Gauss-Newton loss to learn deep features. By performing direct image alignment on the learned features, GN-Net can deliver reliable pose estimation between the images taken from different weather or season conditions. The proposed LM-Net further derives the loss formulation based on Levenberg-Marquardt to improve the robustness against bad initialization compared to the Gauss-Newton method. Inspired by D3VO~\cite{yang2020d3vo}, LM-Reloc also proposes a relative pose estimation network with a correlation layer~\cite{rocco2017convolutional} to regress a pose estimate which is used as the initialization for the optimization.

\noindent{\textbf{Feature-based visual localization.}}
Most approaches for relocalization utilize feature detectors and descriptors, which can either be handcrafted, such as SIFT~\cite{lowe2004distinctive} or ORB~\cite{rublee2011orb}, or especially in the context of drastic lighting and appearance changes can be learned. Recently, many descriptor learning methods have been proposed which follow a \emph{detect-and-describe} paradigm, e.g., SuperPoint~\cite{DeToneCVPRW2018}, D2-Net~\cite{DusmanuCVPR2019}, or R2D2~\cite{revaud2019r2d2}. Moreover, SuperGlue~\cite{sarlin2020superglue}, a learning-based alternative to the matching step of feature-based methods has been proposed and yields significant performance improvements. For a complete relocalization pipeline the local pose refinement part has to be preceded by finding the closest image in a database given a query~\cite{ArandjelovicCVPR2016}. While some approaches~\cite{hloc2018, sarlin2019coarse, taira2018inloc} address the joint problem, in this work, we decouple these two tasks and only focus on the pose refinement part.

\section{Method}\label{methodlm}
In this work, we address the problem of computing the 6DoF pose $\boldsymbol{\xi} \in SE(3)$ between two given images $I$ and $I^{\prime}$. Furthermore, we assume that depths for a sparse set of points $P$ are available, e.g., by running a direct visual SLAM system such as DSO~\cite{dso}.

The overall pipeline of our approach is shown in Figure~\ref{fig:teaser}. It is composed of LM-Net, CorrPoseNet, and a non-linear optimizer using the LM algorithm. LM-Net is trained with a novel loss formulation designed to learn feature descriptors optimally suited for the LM algorithm. The encoder-decoder architecture takes as input a reference image $I$ as well as a target image $I^{\prime}$. The network is trained end-to-end and will produce multi-scale feature maps $F_l$ and $F'_l$, where $l=1,2,3,4$ denotes the different levels of the feature pyramid. In order to obtain an initial pose estimate for the non-linear optimization, we propose Corr\-Pose\-Net, which takes $I$ and $I^{\prime}$ as the inputs and regress their relative pose. Finally, the multi-scale feature maps together with the depths obtained from DSO~\cite{dso} form the non-linear energy function which is minimized using LM algorithm in a coarse-to-fine manner to obtain the final relative pose estimate. In the following, we will describe the individual components of our approach in more detail.

\subsection{Direct Image Alignment with Levenberg-Marquardt}
In order to optimize the pose $\boldsymbol{\xi}$ (consisting of rotation matrix $\mathbf{R}$ and translation $\mathbf{t}$), we minimize the feature-metric error:
\begin{align}
    E(\boldsymbol{\xi}) = \sum_{\mathbf{p} \in P} \bigg \lVert F_l'(\mathbf{p}') - F_l(\mathbf{p}) \bigg \rVert_{\gamma},
\end{align}
where $||\cdot||_\gamma$ is the Huber norm and $\mathbf{p'}$ is the point projected onto the target image $I^{\prime}$ using the depths and the pose: 
\begin{align}
    \mathbf{p}' = \Pi\left(\mathbf{R} \, \Pi^{-1}(\mathbf{p}, d_\mathbf{p}) + \mathbf{t}\right).
\end{align}

This energy function is first minimized on the coarsest pyramid level $1$, whose feature maps $F_1$ have a size of $(w / 8, h / 8)$, yielding a rough pose estimate. The estimate is refined by further minimizing the energy function on the subsequent pyramid levels $2$, $3$, and $4$, where $F_4$ has the size of the original image $(w, h)$. In the following, we provide details of the minimization performed in every level and for simplicity we will denote $F_l$ as $F$ from now on.

Minimization is performed using the Levenberg-Marquardt algorithm. In each iteration we compute the update $\boldsymbol{\delta} \in 
\mathbb{R}^6$ in the Lie algebra $\mathfrak{se}(3)$ as follows:
Using the residual vector $\mathbf{r} \in \mathbb{R}^n$, the Huber weight matrix $\mathbf{W} \in  \mathbb{R}^{n \times n}$, and the Jacobian of the residual vector with respect to the pose $\mathbf{J} \in \mathbb{R}^{n\times 6}$, we compute the Gauss-Newton system:
\begin{align} \label{eq:bigHb}
    \mathbf{H}=\mathbf{J}^T \mathbf{W} \mathbf{J} \text{~and~} \mathbf{b}= -\mathbf{J}^T\mathbf{W}\mathbf{r}.
\end{align}
The damped system can be obtained with either Levenberg's formula~\cite{levenberg1944method}:
\begin{align}
    \mathbf{H'} = \mathbf{H} + \lambda \mathbf{I}
    \label{eq:leven}
\end{align}
or the Marquardt's formula~\cite{marquardt1963algorithm}:
\begin{align}
    \mathbf{H'} = \mathbf{H} + \lambda \, \text{diag}(\mathbf{H})
    \label{eq:marq}
\end{align}
depending on the specific application.

The update $\boldsymbol{\delta}$ and the pose $\boldsymbol{{\xi}}^i$ in the iteration $i$ are computed as:
\begin{align}
    \boldsymbol{\delta} = \mathbf{H'}^{-1} \mathbf{b} \text{~and~} \boldsymbol{{\xi}}^i = \boldsymbol{\delta} \boxplus \boldsymbol{{\xi}}^{i-1},
\end{align}
where $\boxplus: \mathfrak{se}(3) \times SE(3) \rightarrow SE(3)$ is defined as in~\cite{dso}.

The parameter $\lambda$ can be seen as an interpolation factor between gradient descent and the Gauss-Newton algorithm. When $\lambda$ is high the method behaves like gradient descent with a small step size, and when it is low it is equivalent to the Gauss-Newton algorithm.
In practice, we start with a relatively large $\lambda$ and multiply it by $0.5$ after a successful iteration, and by $4$ after a failed iteration~\cite{dso}.

\begin{figure}[t]
    \centering
    \includegraphics[width=\linewidth]{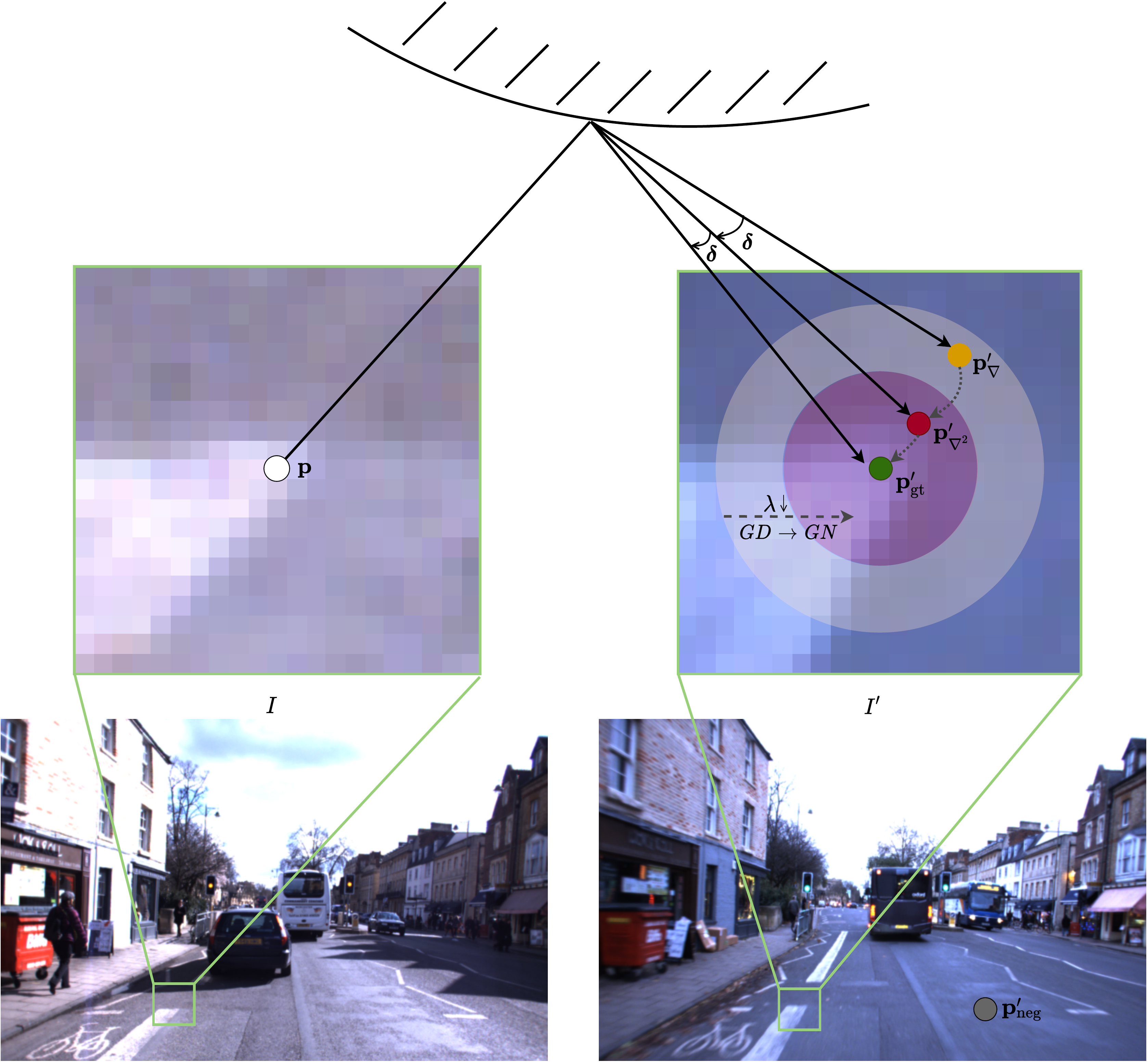}
    \caption{
    Visualization of the typical behavior of direct image alignment with Levenberg-Marquardt. Initially, the projected point position (orange point, $\mathbf{p}_\nabla'$) is far away from the correct solution (green point, $\mathbf{p}_\text{gt}'$), and $\lambda$ is large, yielding an update step similar to gradient descent. After some iterations the projected point position gets closer to the optimum (red point, $\mathbf{p}_{\nabla^2}'$) and at the same time $\lambda$ will get smaller, leading to an update step similar to the Gauss-Newton algorithm.
This is the intuition behind our point sampling strategy, where we utilize the ground-truth correspondence $\mathbf{p}_\text{gt}'$ for Equation (\ref{eq:firstloss}), a negative $\mathbf{p}_\text{neg}'$ sampled across the whole image for Equation (\ref{eq:secondloss}), a negative $\mathbf{p}_\nabla'$ sampled in a far vicinity for Equation (\ref{eq:gdloss}), and a negative $\mathbf{p}_{\nabla^2}'$ sampled in a close vicinity for Equation (\ref{eq:gnloss}).
}
    \label{fig:lmprogress}
\end{figure}

Figure~\ref{fig:lmprogress} shows the typical behaviour of the algorithm.
In the beginning the initial pose is inaccurate, resulting in projected point positions, which are a couple of pixels away from the correct location. $\lambda$ will be high meaning that the algorithm will behave similar to gradient descent. After a couple of iterations, the pose got more accurate, and the projected points are in a closer vicinity to the correct location. By now, $\lambda$ has probably decreased, so the algorithm will behave more similar to the Gauss-Newton algorithm. Now we expect the algorithm to converge quickly.

\subsection{Loss Formulation for Levenberg-Marquardt} \label{sec:lossformulation}

The key contribution of this work is LM-Net which provides feature maps $F$ that improve the convergence behaviour of the LM algorithm and, in the meantime, are invariant to different conditions. We train our network in a Siamese fashion based on ground-truth pixel correspondences.

In this section, $\mathbf{p}$ denotes a reference point (located on image $I$) and the ground-truth correspondence (located on image $I^{\prime}$) is $\mathbf{p}_{\text{gt}}'$. For the loss functions explained below we further categorize $\mathbf{p}'$ into $\mathbf{p}_\text{neg}'$, $\mathbf{p}_\nabla'$, and $\mathbf{p}_{\nabla^2}'$, which is realized by using different negative correspondence sampling. Our loss formulation is inspired by the typical behaviour of the Levenberg-Marquardt algorithm explained in the previous section (see Figure~\ref{fig:lmprogress}). For a point, we distinguish four cases which can happen during the optimization:

\begin{enumerate}
    \item The point is at the correct location ($\mathbf{p}_{\text{gt}}'$).
    \item The point is an outlier ($\mathbf{p}_\text{neg}'$).
    \item The point is relatively far from the correct solution ($\mathbf{p}_\nabla'$).
    \item The point is very close to the correct solution ($\mathbf{p}_{\nabla^2}'$).
\end{enumerate}

In the following we will derive a loss function for each of the 4 cases:\\

\noindent{\textbf{1. The point is already at the correct location.}}
In this case we would like the residual to be as small as possible, in the best case 0.
\begin{align}
E_\text{pos} = \lVert F'(\mathbf{p}_{\text{gt}}') - F(\mathbf{p}) \rVert^2
\label{eq:firstloss}
\end{align}
\noindent{\textbf{2. The point is an outlier or the pose estimate is completely wrong.}} 
In this case the projected point position can be at a completely different location than the correct correspondence. In this scenario we would like the residual of this pixel to be very large to reflect this, and potentially reject a wrong update.
To enforce this property we sample a negative correspondences $\mathbf{p}_\text{neg}'$ uniformly across the whole image, and compute
\begin{align}
    E_\text{neg} = \max \left( M - \lVert F'(\mathbf{p}_\text{neg}') - F(\mathbf{p}) \rVert^2, 0\right)
     \label{eq:secondloss}
\end{align}
where $M$ is the margin how large we would like the energy of a wrong correspondence to be. In practice, we set it to $1$. 

\noindent{\textbf{3. The predicted pose is relatively far away from the optimum,}} meaning that the projected point position will be a couple of pixels away from the correct location. As this typically happens during the beginning of the optimization we assume that $\lambda$ will be relatively large and the algorithm behaves similar to gradient descent.
In this case we want that the gradient of this point is oriented in the direction of the correct solution, so that the point has a positive influence on the update step. 

For computing a loss function to enforce this property we sample a random negative correspondence $\mathbf{p}_\nabla'$ in a relatively large vicinity around the correct solution (in our experiments we use 5 pixels distance).
Starting from this negative correspondence $\mathbf{p}_\nabla'$ we first compute the $2 \times 2$ Gauss-Newton system for this individual point, similarly to how it is done for optical flow estimation using Lucas-Kanade:
\begin{align} \label{eq:2dres}
    \mathbf{r}_\mathbf{p}(\mathbf{p}, \mathbf{p}_\nabla') = \mathbf{F'}(\mathbf{p}_\nabla') - \mathbf{F}(\mathbf{p}) 
\end{align}
\begin{align} \label{eq:2djac}
\mathbf{J}_\mathbf{p} = \frac{d \mathbf{F'}(\mathbf{p}_\nabla')}{d\mathbf{p}_\nabla'} \text{ and } \mathbf{H}_\mathbf{p} = \mathbf{J}_\mathbf{p}^T \mathbf{J}_\mathbf{p} \text{ and } \mathbf{b}_\mathbf{p} = \mathbf{J}_\mathbf{p}^T \mathbf{r}_\mathbf{p}
\end{align}

We compute the damped system using a relatively large fixed $\lambda_f$, as well as the optical flow step\footnote{Here we use Equation (\ref{eq:leven}) instead of Equation (\ref{eq:marq}) since we find it more stable for training LM-Net.}
\begin{align}
    \mathbf{H}_\mathbf{p}' = \mathbf{H}_\mathbf{p} + \lambda_f \, \mathbf{I} \text{ and } \mathbf{p}_{\text{after}}' = \mathbf{p}_\nabla' + \mathbf{H}_\mathbf{p}'^{-1} \mathbf{b}_\mathbf{p}.
\end{align}

In order for this point to have a useful contribution to the direct image alignment, this update step should move in the correct direction by at least $\delta$. We enforce this using a Gradient-Descent loss function which is small only if the distance to the correct correspondence \emph{after} the update is smaller than before the update:
\begin{align}
   E_{\text{GD}} =  \max\left(\lVert \mathbf{p}_{\text{after}}' - \mathbf{p}_{\text{gt}}' \rVert^2 - \lVert \mathbf{p}_\nabla' - \mathbf{p}_{\text{gt}}' \rVert^2 + \delta, 0\right)
   \label{eq:gdloss}
\end{align}
In practice, we choose $\lambda_f=2.0$ and $\delta=0.1$.

\noindent{\textbf{4. The predicted pose is very close to the optimum,}} yielding a projected point position in very close proximity of the correct correspondence, and typically $\lambda$ will be very small, so the update will mostly be a Gauss-Newton step. In this case we would like the algorithm to converge as quickly as possible, with subpixel accuracy.
We enforce this using the Gauss-Newton loss~\cite{gnnet}.
To compute it we first sample a random negative correspondence $\mathbf{p}_{\nabla^2}'$ in a 1-pixel vicinity around the correct location. Then we use Equations (\ref{eq:2dres}) and (\ref{eq:2djac}), replacing $\mathbf{p}_{\nabla}'$ with $\mathbf{p}_{\nabla^2}'$ to obtain the Gauss-Newton system formed by $\mathbf{H}_\mathbf{p}$ and $\mathbf{b}_\mathbf{p}$. We compute the updated pixel location:

\begin{align}
    \mathbf{p}_{\text{after}}' = \mathbf{p}_{\nabla^2}' + (\mathbf{H}_\mathbf{p} + \epsilon \, \mathbf{I})^{-1} \mathbf{b}_\mathbf{p}
\end{align}

Note that in contrast to the computation of the LM-Loss (Equation (\ref{eq:gdloss})), in this case $\epsilon$ is just added to ensure invertibility and therefore $\epsilon$ is much smaller than the $\lambda_f$ used above.
The Gauss-Newton loss is computed with:
\begin{align}
     E_\text{GN} = \frac{1}{2}(\mathbf{p}_{\text{after}}' - \mathbf{p}_{\text{gt}}')^T \mathbf{H}_\mathbf{p} (\mathbf{p}_{\text{after}}' - \mathbf{p}_{\text{gt}}') \nonumber \\ + \log(2\pi) - \frac{1}{2} \log(|\mathbf{H}_\mathbf{p}|) \label{eq:gnloss}
\end{align}

Note how all our 4 loss components use a different way to sample the involved points, depicted also in Figure \ref{fig:lmprogress}. With the derivation above we argue that each loss component is important to achieve optimal performance and we demonstrate this in the results section. Note that the Gauss-Newton systems computed for the GD-Loss and the GN-Loss are very relevant for the application of direct image alignment. In fact the full Gauss-Newton system containing all points (Equation (\ref{eq:bigHb})), can be computed from these individual Gauss-Newton systems (Equation (\ref{eq:2djac})) by simply summing them up and multiplying them with the derivative with respect to the pose~\cite{gnnet}.

\subsection{CorrPoseNet}

In order to deal with the large baselines between the images, we propose CorrPoseNet to regress the relative pose between two images $I$ and $I^{\prime}$, which serves as the initialization of LM optimization.
As our network shall work even in cases of large baselines and strong rotations, we utilize the correlation layer proposed in~\cite{rocco2017convolutional} which is known to boost the performance of affine image transformation and optical flow~\cite{dgcnet} estimation for large displacements, but has not been applied to pose estimation before.

Our network first computes deep features $\mathbf{f}_\text{corr}$, $\mathbf{f}_\text{corr}' \in \mathbb{R}^{h\times w \times c}$ from both images individually using multiple strided convolutions with ReLU activations in between. Then the correlation layer correlates each pixel from the normalized source features with each pixel from the normalized target features yielding the correlation map $\mathbf{c} \in \mathbb{R}^{h \times w \times (h \times w)}$:
\begin{align}
    \mathbf{c}(i, j, (i', j')) = \mathbf{f}_\text{corr}(i, j)^T \mathbf{f}_\text{corr}'(i', j')
\end{align}
The correlation map is then normalized in the channel dimension and fed into 2 convolutional layers each followed by batch norm and ReLU. Finally we regress the Euler angle $\mathbf{r}^\text{euler}$ and translation $\mathbf{t}$ using a fully connected layer.
More details on the architecture are shown in the supplementary material.

We train CorrPoseNet from scratch with image pairs and groundtruth poses $\mathbf{r}^\text{euler}_\text{gt}, \mathbf{t}_\text{gt}$. We utilize an L2-loss working directly on Euler angles and translation:
\begin{align}
    E = \lVert \mathbf{t} - \mathbf{t}_\text{gt} \rVert_2 + \lambda 
    \lVert \mathbf{r}^\text{euler} - \mathbf{r}^\text{euler}_\text{gt} \rVert_2,
\end{align}
where $\lambda$ is the weight, which we set to $10$ in practice.

As the distribution of groundtruth poses in the Oxford training data is limited we apply the following data augmentation. 
We first generate dense depths for all training images using a state-of-the-art dense stereo matching algorithm~\cite{ganet}. The resulting depths are then used to warp the images to a different pose sampled from a uniform distribution. In detail, we first warp the depth image to the random target pose, then inpaint the depth image using the OpenCV implementation of Navier Stokes, and finally warp our image to the target pose using this depth map. Note that the dense depths are only necessary for training, not for evaluation. We show an ablation study on the usage of correlation layers and the proposed data augmentation in the supplementary material. 

\section{Experiments}

\begin{figure*}[t]
  \centering
    \begin{subfigure}{.49\linewidth}
    \includegraphics[width=\linewidth]{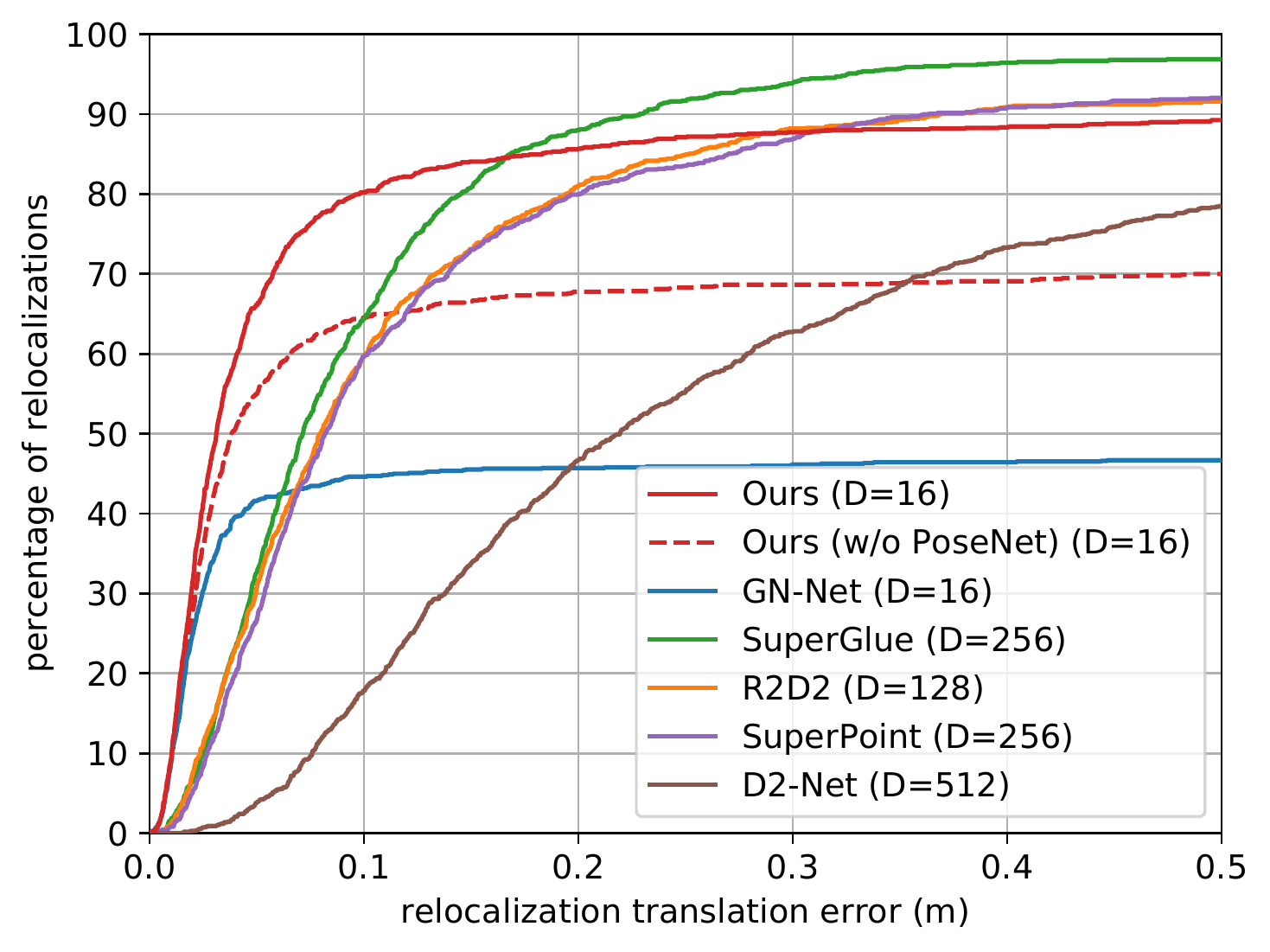}
    \caption{Translation error.}
    \label{fig:carlatranslation}
  \end{subfigure}
  \begin{subfigure}{.49\linewidth}
    \includegraphics[width=\linewidth]{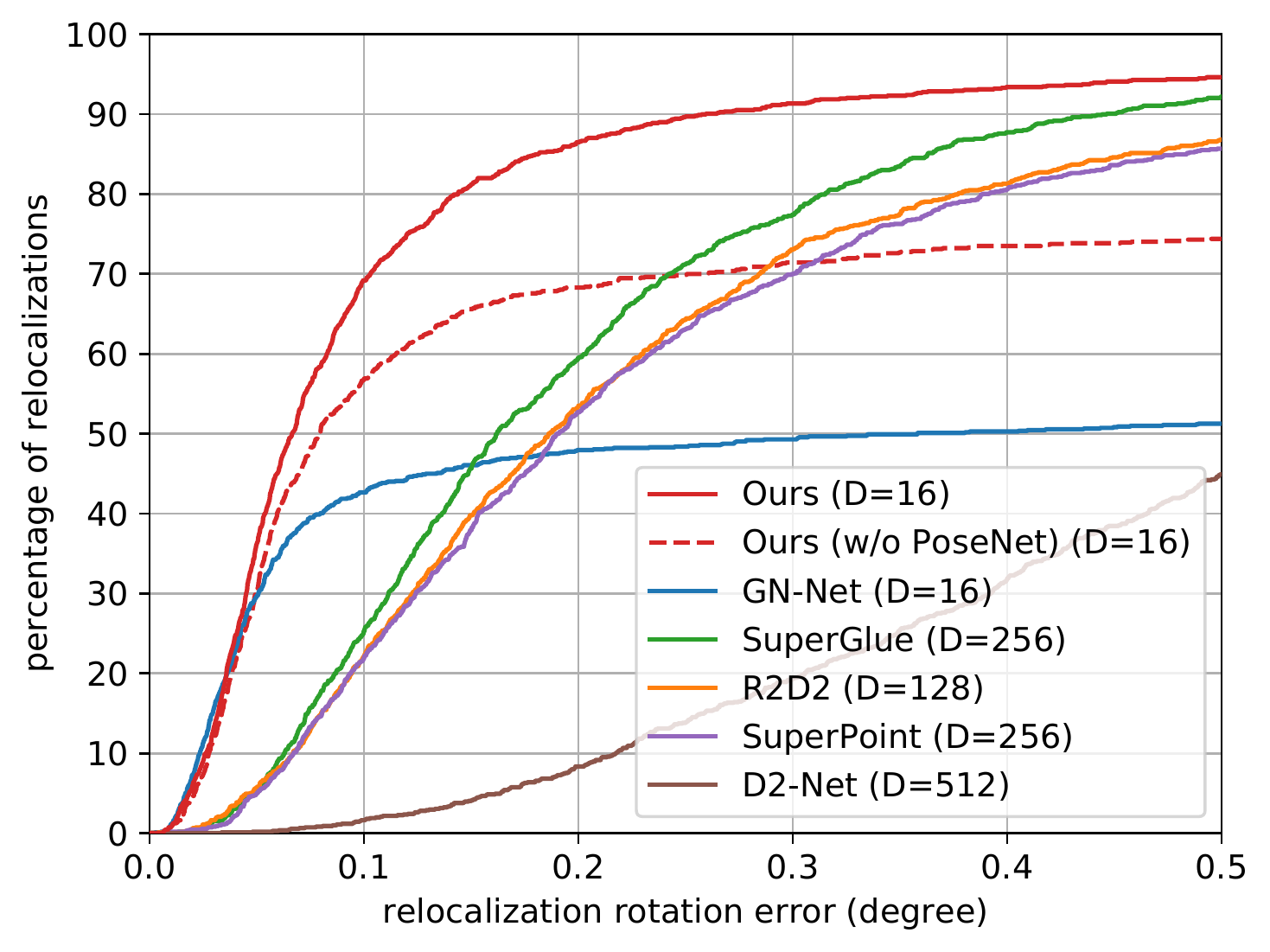}
    \caption{Rotation error.}
    \label{fig:carlarotation}
  \end{subfigure}
  \caption{Results on the CARLA relocalization tracking benchmark test data~\cite{gnnet}.
  For each error threshold we show the percentage of relocalizations (cumulative error plot) for LM-Reloc (ours) and other state-of-the-art methods. Compared to the indirect methods our approach exhibits significantly better accuracy in both translation and rotation, while having a similar robustness. Compared to GN-Net, the novel loss formulation (see red dashed line), and the CorrPoseNet (see red line) both boost the robustness. $D$ is the feature dimensionality.
  }
  \label{fig:carla_test}
\end{figure*}

We evaluate our method on the relocalization tracking benchmark proposed in~\cite{gnnet}, which contains images created with the CARLA simulator~\cite{DosovitskiyCoRL2017}, and scenes from the Oxford RobotCar dataset~\cite{MaddernIJRR2017}. We train our method on the respective datasets from scratch. LM-Net is trained using the Adam optimizer with a learning rate of $10^{-6}$ and for CorrPose\-Net we use a learning rate of $10^{-4}$. For both networks we choose hyperparameters and epoch based on the results on the validation data. Our networks use the same hyperparameters for all experiments except where stated otherwise; the direct image alignment code is slightly adapted for Oxford RobotCar, mainly to improve performance when the ego-vehicle is standing still.

As the original relocalization tracking benchmark~\cite{gnnet} does not include validation data on Oxford RobotCar we have manually aligned two new sequences, namely \emph{2015-04-17-09-06-25} and \emph{2015-05-19-14-06-38}, and extend the benchmark with these sequences as validation data.

\noindent{\textbf{Evaluation metrics:}} We evaluate the predicted translation $\mathbf{t}_{\text{est}}$ and rotation $\mathbf{R}_{\text{est}}$ against the ground-truth $\mathbf{t}_{\text{gt}}$ and $\mathbf{R}_{\text{gt}}$ according to Equations (\ref{eq:trans_error}) and (\ref{eq:rot_err}). 

\begin{align}
    t_{\Delta} &= \lVert \mathbf{t}_{\text{est}} - \mathbf{t}_{\text{gt}} \rVert_2 \label{eq:trans_error} \\
    R_{\Delta} &= \text{arccos}\left(\frac{\text{trace}(\mathbf{R}_{\text{est}}^{-1} \mathbf{R}_{\text{gt}}) - 1} {2}\right) \label{eq:rot_err}
\end{align}

In this section, we plot the cumulative translation and rotation error until $0.5$m and $0.5^{\circ}$, respectively. For quantitative results we compute the area under curve (AUC) of these cumulative curves in percent, which we denote as $t_\text{AUC}$ for translation and $R_\text{AUC}$ for rotation from now on.

We evaluate the following direct methods:

\noindent{\textbf{Ours:}}
The full LM-Reloc approach consisting of Corr\-Pose\-Net, LM-Net features and direct image alignment based on Levenberg-Marquardt. The depths used for the image alignment are estimated with the stereo version \cite{wang2017stereoDSO} of DSO \cite{dso}. 

\noindent{\textbf{Ours (w/o CorrPoseNet):}}
For a more fair comparison to GN-Net we use identity as initialization for the direct image alignment instead of CorrPoseNet. This enables a direct comparison between the two loss formulations.

\noindent{\textbf{GN-Net~\cite{gnnet}:}}
In this work, we have also improved the parameters of the direct image alignment pipeline based on DSO~\cite{dso}. Thus we have re-evaluated GN-Net with this improved pipeline to make the comparison as fair as possible. These re-evaluated results are better than the results computed in the original GN-Net paper.

\noindent{\textbf{Baseline methods:}}
Additionally, we evaluate against current state-of-the-art indirect methods, namely SuperGlue~\cite{sarlin2020superglue}, R2D2~\cite{revaud2019r2d2}, SuperPoint~\cite{DeToneCVPRW2018}, and D2-Net~\cite{DusmanuCVPR2019}. For these methods, we estimate the relative pose using the models provided by the authors and the OpenCV implementation of solvePnPRansac. We have tuned the parameters of RANSAC on the validation data and used $1000$ iterations and a reprojection error threshold of $3$ for all methods. For estimating depth values at keypoint locations we use OpenCV stereo matching. It would be possible to achieve a higher accuracy by using SfM and MVS solutions such as COLMAP~\cite{schoenberger2016sfm}. However, one important disadvantage of these approaches is, that building a map is rather time consuming and computationally expensive, whereas all other approaches evaluated on the benchmark~\cite{gnnet} are able to create the map close to real-time, enabling applications like long-term loop-closure and map-merging.

\subsection{CARLA Relocalization Benchmark}

\begin{table}[t]
\caption{This table shows the AUC until $0.5$ meters / $0.5$ degrees for the relocalization error on the CARLA relocalization tracking benchmark test data. Powered by our novel loss formulation and the combination with CorrPoseNet, LM-Reloc achieves lower rotation and translation errors compared to the state-of-the-art.}
\label{tab:carla_test}
\begin{center}
\begin{tabular}{c | c c}
\hline
     Method & $t_{\text{AUC}}$ & $R_{\text{AUC}}$ \\
     \hline
     Ours & \textbf{80.65} & \textbf{77.83} \\
     SuperGlue~\cite{sarlin2020superglue} & 78.99 & 59.31 \\
     R2D2~\cite{revaud2019r2d2} & 73.47 & 54.42 \\
     SuperPoint~\cite{DeToneCVPRW2018} & 72.76 & 53.38 \\
     D2-Net~\cite{DusmanuCVPR2019} & 47.62 & 16.47 \\
     \hline
     Ours (w/o CorrPoseNet) & 63.88 & 61.9 \\
     GN-Net~\cite{gnnet} & 43.72 & 44.08 \\
\hline
\end{tabular}
\end{center}
\end{table}

\begin{table*}[t]
\caption{Results on the Oxford RobotCar relocalization tracking benchmark~\cite{gnnet}. We compare LM-Net (Ours) against other state-of-the-art methods (SuperGlue, R2D2, SuperPoint, and D2-Net). As can be seen from the results, our method almost consistently outperforms other SOTA approaches in terms of rotation AUC whilst achieving comparable results on translation AUC.}
\label{tab:oxfordtest}
\begin{center}
  \begin{tabular}{c | c c | c c | c c | c c | c c}
    \hline
Sequence & \multicolumn{2}{|c}{Ours} & \multicolumn{2}{|c}{SuperGlue~\cite{sarlin2020superglue}} & \multicolumn{2}{|c}{R2D2~\cite{revaud2019r2d2}} & \multicolumn{2}{|c}{SuperPoint~\cite{DeToneCVPRW2018}} & \multicolumn{2}{|c}{D2-Net~\cite{DusmanuCVPR2019}} \\
 & $t_{\text{AUC}}$ & $R_{\text{AUC}}$ & $t_{\text{AUC}}$ & $R_{\text{AUC}}$ & $t_{\text{AUC}}$ & $R_{\text{AUC}}$ & $t_{\text{AUC}}$ & $R_{\text{AUC}}$ & $t_{\text{AUC}}$ & $R_{\text{AUC}}$ \\
\hline
Sunny-Overcast & 79.83 & \textbf{55.48} & \textbf{81.01} & 52.83 & 80.86 & 53.57 & 78.95 & 50.03 & 71.93 & 39.0 \\
Sunny-Rainy & 71.54 & \textbf{43.7} & \textbf{75.58} & 40.59 & 74.84 & 41.23 & 69.76 & 37.12 & 65.63 & 27.5 \\
Sunny-Snowy & 59.69 & \textbf{44.06} & \textbf{63.57} & 43.64 & 62.92 & 41.78 & 60.85 & 40.02 & 55.65 & 30.86 \\
Overcast-Rainy & 80.54 & \textbf{63.7} & 79.99 & 61.64 & \textbf{81.29} & 61.23 & 80.36 & 61.56 & 75.66 & 51.06 \\
Overcast-Snowy & 55.38 & 47.88 & 57.67 & 47.16 & \textbf{57.68} & \textbf{48.41} & 55.39 & 44.96 & 51.17 & 34.54 \\
Rainy-Snowy & 68.57 & \textbf{41.67} & 69.91 & 39.87 & \textbf{71.79} & 39.86 & 67.7 & 38.05 & 61.91 & 27.74 \\
\hline
\end{tabular}
\end{center} 
\end{table*}

\begin{table}[t]
\caption{This table shows the results on the Oxford RobotCar relocalization tracking benchmark test data against GN-Net. Thanks to our LM-based loss formulation we consistently outperform GN-Net on all sequences.}
\label{tab:oxfordtestnoposenet}
\begin{center}
\resizebox{\columnwidth}{!}{%
\begin{tabular}{c | c c | c c}
\hline
     Sequence & \multicolumn{2}{|c}{Ours (w/o CorrPoseNet)} & \multicolumn{2}{|c}{GN-Net~\cite{gnnet}} \\
     & $t_{\text{AUC}}$ & $R_{\text{AUC}}$ & $t_{\text{AUC}}$ & $R_{\text{AUC}}$ \\
     \hline
Sunny-Overcast & \textbf{79.61} & \textbf{55.45} & 73.53 & 49.31 \\
Sunny-Rainy & \textbf{70.46} & \textbf{42.86} & 64.58 & 37.27 \\
Sunny-Snowy & \textbf{59.7} & \textbf{44.17} & 55.27 & 41.36 \\
Overcast-Rainy & \textbf{79.67} & \textbf{63.08} & 75.72 & 60.13 \\
Overcast-Snowy & \textbf{54.94} & \textbf{47.19} & 51.34 & 42.91 \\
Rainy-Snowy & \textbf{66.23} & \textbf{39.93} & 62.63 & 36.2 \\
\hline
\end{tabular}
}
\end{center}
\end{table}

Figure~\ref{fig:carla_test} depicts the results on the test data of the CARLA benchmark. For all methods we show the cumulative error plot for translation in meters and rotation in degree. It can be seen that our method is more accurate than the state-of-the-art while performing similarly in terms of robustness.
We also show the AUC for the two Figures in Table \ref{tab:carla_test}. 
Compared to GN-Net it can be seen that our new loss formulation significantly improves the results, even when used without the CorrPoseNet as initialization.
The figure conveys that the direct methods (Ours, GN-Net) are more accurate than the evaluated indirect methods.

\subsection{Oxford RobotCar Relocalization Benchmark} 

We compare to the state-of-the-art indirect methods on the 6 test sequence pairs consisting of the sequences \emph{2015-02-24-12-32-19} (sunny), \emph{2015-03-17-11-08-44} (overcast), \emph{2014-12-05-11-09-10} (rainy), and \emph{2015-02-03-08-45-10} (snowy).
In Table~\ref{tab:oxfordtest}, we show the area under curve until $0.5$ meters / $0.5$ degrees for all methods. It can be seen that our method clearly outperforms the state-of-the-art in terms of rotation accuracy, while being competitive in terms of translation error.
It should be noted that the ground-truth for these sequences was generated using ICP alignment of the 2D-LiDAR data accumulated for 60 meters. We have computed that the average root mean square error of the ICP alignment is 16 centimeters. Therefore, especially the ground-truth translations have limited accuracy. As can be seen from Figure~\ref{fig:carla_test}, the accuracy improvements our method provides are especially visible in the range below $0.15$ meters which is hard to measure on this dataset.
The rotation error of LiDAR alignment is lower than the translational one, which is why we clearly observe the improvements of our method on the rotations.

In Table~\ref{tab:oxfordtestnoposenet}, we compare LM-Net without the CorrPoseNet to GN-Net. Due to our novel loss formulation LM-Net outperforms the competitor on all sequences significantly.

\subsection{Ablation Studies}

\begin{figure}
  \centering
    \includegraphics[width=\linewidth]{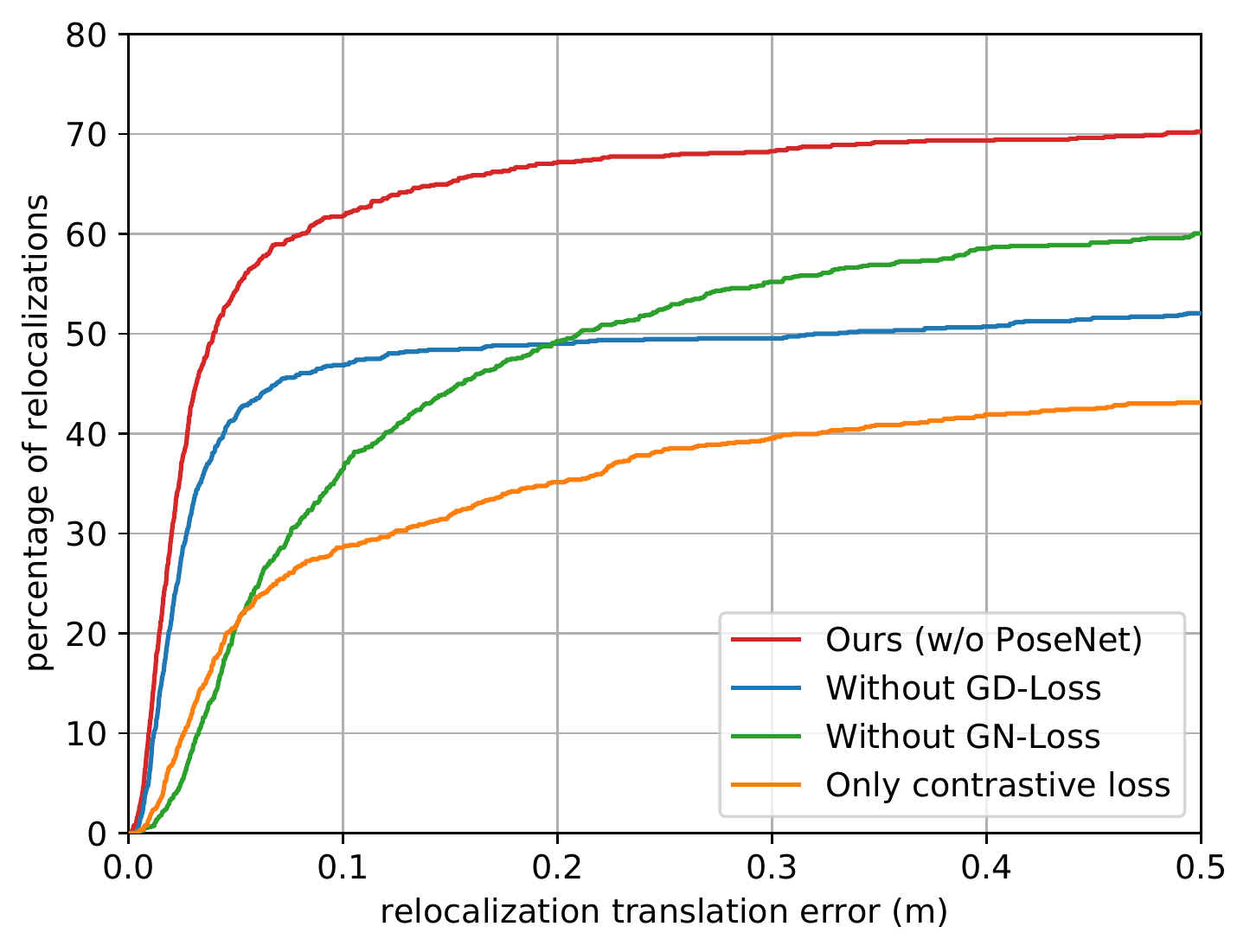}
  \caption{This plot shows our ablation study for removing different loss parts on the CARLA relocalization tracking benchmark. Without the GD-loss the achieved robustness is reduced, whereas removing the GN-loss leads to decreased accuracy. Using our full loss formulation yields a large improvement.}
  \label{fig:carla_ablation}
\end{figure}
We evaluate LM-Net on the CARLA validation data with and without the various losses (Figure \ref{fig:carla_ablation}). Compared to a normal contrastive loss, the given loss formulation is a large improvement. As expected, $E_\text{GD}$ (green line) mainly improves the robustness, whereas $E_\text{GN}$ (blue line) improves the accuracy. Only when used together (our method) we achieve large robustness and large accuracy, confirming our theoretical derivation in Section~\ref{methodlm}.

\begin{figure*}[t]
  \centering
  \includegraphics[width=1.0\linewidth]{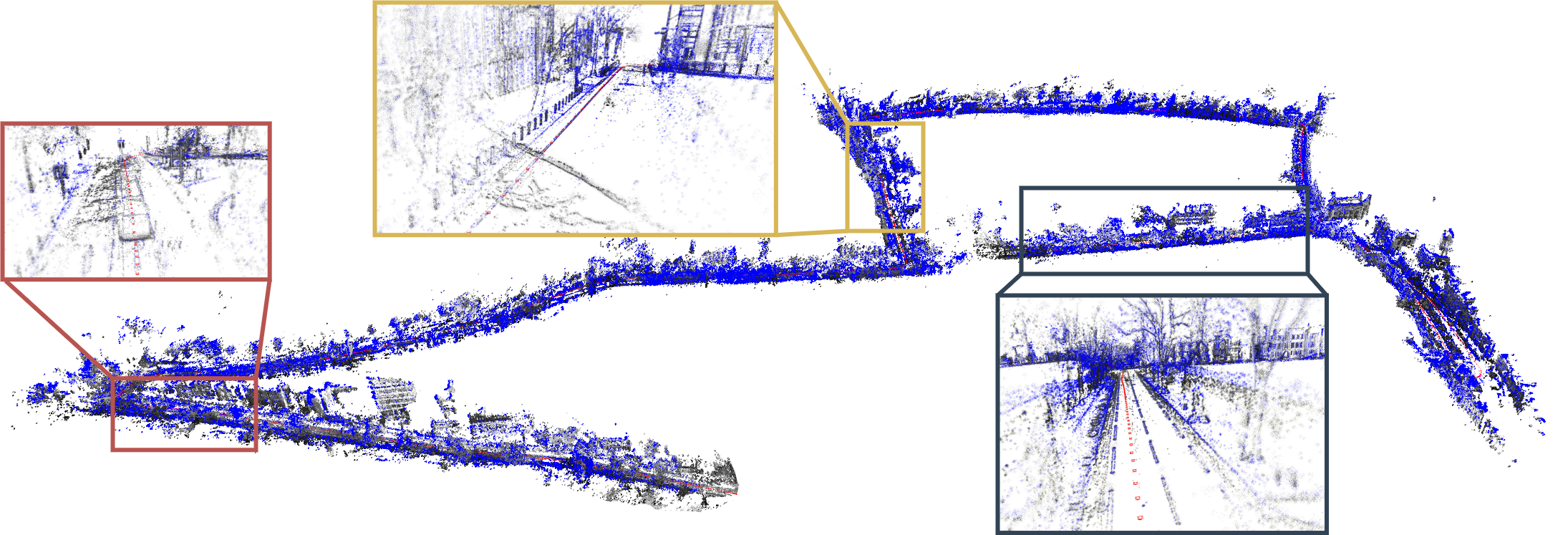}
  \caption{
  This figure shows a point cloud from a sunny reference map (grey points) overlayed with the point cloud from a relocalized snowy sequence (blue points). The well aligned point clouds demonstrate the high relocalization accuracy of LM-Reloc.
  }
  \label{fig:qualitative}
\end{figure*}

\begin{figure}[t]
    \centering
    \includegraphics[width=\linewidth]{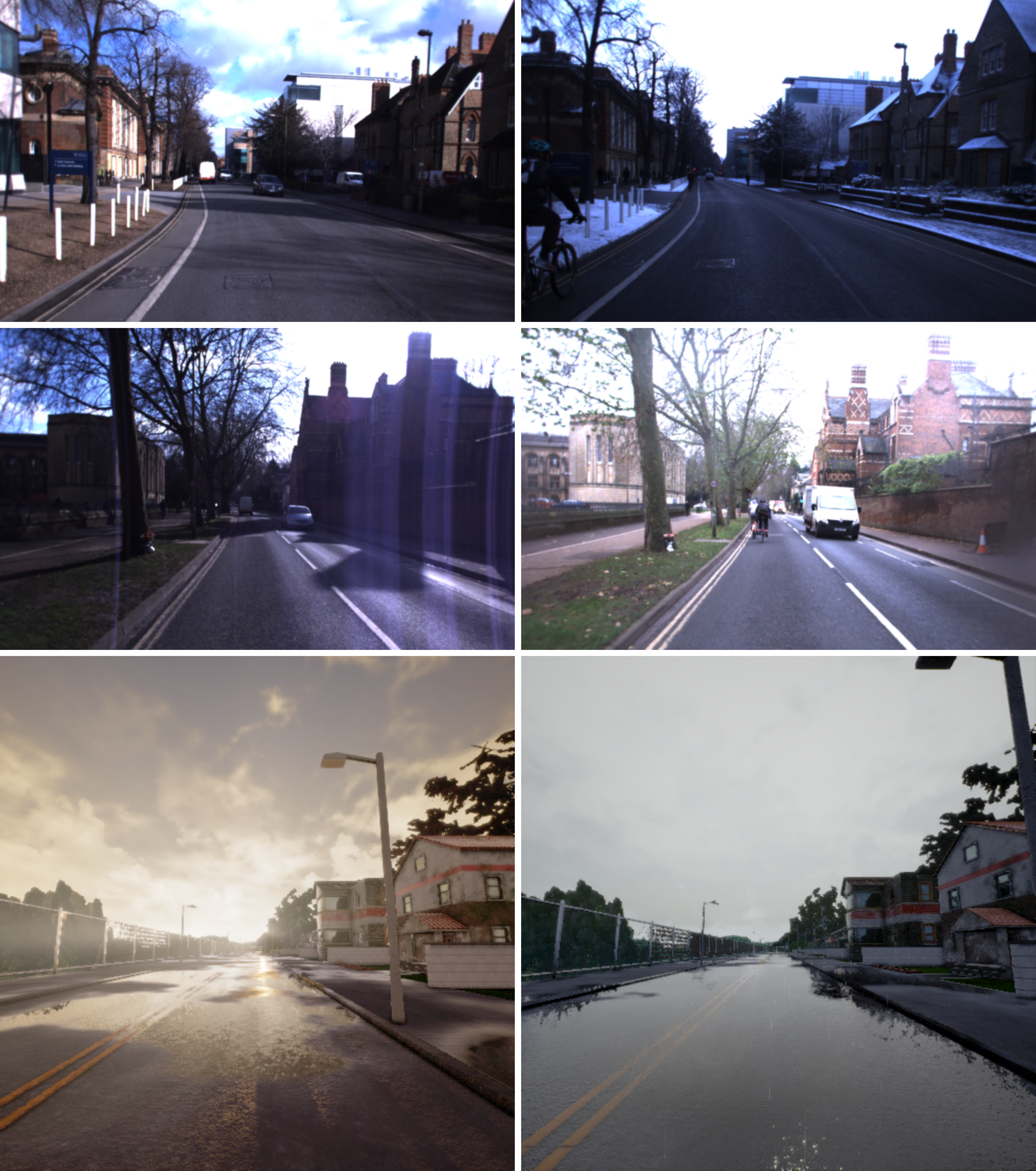}
    \caption{Example image pairs from the relocalization tracking benchmark which have been successfully relocalized by LM-Reloc (with an accuracy of better than $10$ cm). Top row: Oxford sunny against snowy condition, middle row: Oxford sunny against rainy condition, bottom row: CARLA benchmark.
}
    \label{fig:examples}
\end{figure}

\subsection{Qualitative Results}

To demonstrate the accuracy of our approach in practice, we show qualitative results on the Oxford RobotCar dataset.
We track the snowy test sequence \emph{2015-02-03-08-45-10} using Stereo DSO~\cite{wang2017stereoDSO} and at the same time perform relocalization against the sunny reference map \emph{2015-02-24-12-32-19}.
Relocalization between the current keyframe and the closest map image is performed using LM-Net. Initially, we give the algorithm the first corresponding map image (which would in practice be provided by an image retrieval approach such as NetVLAD~\cite{ArandjelovicCVPR2016}). 
Afterwards we find the closest map image for each keyframe using the previous solution for the transformation between the map and the current SLAM world $T_{w\_m}$.
We visualize the current point cloud (blue) and the point cloud from the map (grey) overlayed using the smoothed $T_{w\_m}$ (Figure~\ref{fig:qualitative}). The point clouds will align only if the relocalization is accurate.
As can be seen in Figure~\ref{fig:qualitative}, the lane markings, poles, and buildings between the reference and query map align well, hence qualitatively showing the high relocalization accuracy of our method. We recommend watching the video at~\url{https://vision.in.tum.de/lm-reloc}. In Figure~\ref{fig:examples} we show example images from the benchmark. 

\section{Conclusion}
We have presented LM-Reloc as a novel approach for direct visual localization. In order to estimate the relative 6DoF pose between two images from different conditions, our approach performs direct image alignment on the trained features from LM-Net without relying on feature matching or RANSAC. In particular, with the loss function designed seamlessly for the Levenberg-Marquart algorithm, LM-Net provides deep feature maps that coin the characteristics of direct image alignment and are also invariant to changes in lighting and appearance of the scene. The experiments on the CARLA and Oxford RobotCar relocalization tracking benchmark exhibit the state-of-the-art performance of our approach. In addition, the ablation studies also show the effectiveness of the different components of LM-Reloc.

\clearpage
{\small
\bibliographystyle{ieee}
\bibliography{main}
}

% supplement
\clearpage


\section*{Supplementary material}

# LM-Reloc: Levenberg-Marquardt Based Direct Visual Relocalization: Supplementary Material

Lukas von Stumberg[1,2*]   Patrick Wenzel[1,2*]   Nan Yang[1,2]   Daniel Cremers[1,2]
[1] Technical University of Munich   [2] Artisense

## 1. Video

As mentioned in the paper, we provide a video of the qualitative relocalization demo, which is available at https://vision.in.tum.de/lm-reloc.

## 2. Network Architecture

**CorrPoseNet.** The CorrPoseNet takes 2 images ($I$ and $I'$) as the input and outputs the relative pose $R, t$ between those images. The overall network architecture of the CorrPoseNet is depicted in Figure 1. The convolutional blocks consist of in total 9 convolutional layers followed by ReLU activations. The architectural details of the convolutional blocks are listed in Table 1. The correlation layer which takes the output of the convolutional blocks as input is described in the main paper. The correlation layer is followed by the regression block which regresses the relative pose. The layers of the regression block are listed in Table 2. The output of the network is the rotation $R$ as Euler angles and translation $t$.

**Table 1:** Network architecture and parameters of the convolutional blocks. **k** denotes kernel size, **s** stride, and **p** padding.

| Convolutional blocks | | | | | | |
| --- | --- | --- | --- | --- | --- | --- |
| layer | in-chns | out-chns | k | s | p | activation |
| conv0 | 3 | 16 | 16 | 2 | 3 | ReLU |
| conv1 | 16 | 32 | 5 | 2 | 2 | ReLU |
| conv2 | 32 | 64 | 3 | 2 | 1 | ReLU |
| conv3 | 64 | 64 | 3 | 1 | 0 | ReLU |
| conv4 | 64 | 128 | 3 | 2 | 2 | ReLU |
| conv5 | 128 | 128 | 3 | 1 | 1 | ReLU |
| conv6 | 128 | 256 | 3 | 2 | 1 | ReLU |
| conv7 | 256 | 256 | 3 | 1 | 1 | ReLU |

**LM-Net.** We adopt U-Net [1] as the encoder of LM-Net. However, we change the decoder part of the architecture in the following way. Starting from the coarsest level, we upsample (with bilinear interpolation) the feature maps by 2 and concatenate those feature maps with the feature map of the higher level. This is followed by $1 \times 1$ convolutional

*Equal contribution.

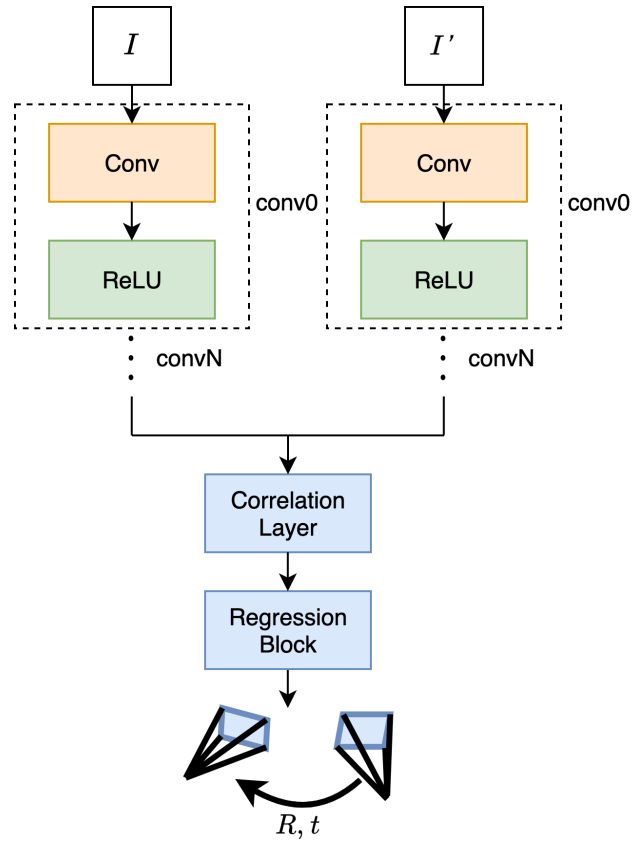

**Figure 1:** Network architecture of CorrPoseNet.

filters. This procedure is repeated 4 times. This results in the feature pyramid maps as described in Table 3.

## 3. Ablation Study Correlation Layer

We demonstrate the impact of the Correlation layer in the proposed CorrPoseNet. We compare it to a simpler pose estimation network where the correlation and regression layers are replaced with two $1 \times 1$ convolutions with 3 output-channels each, which directly regress rotation and Euler angles. This simpler PoseNet has one more convo-

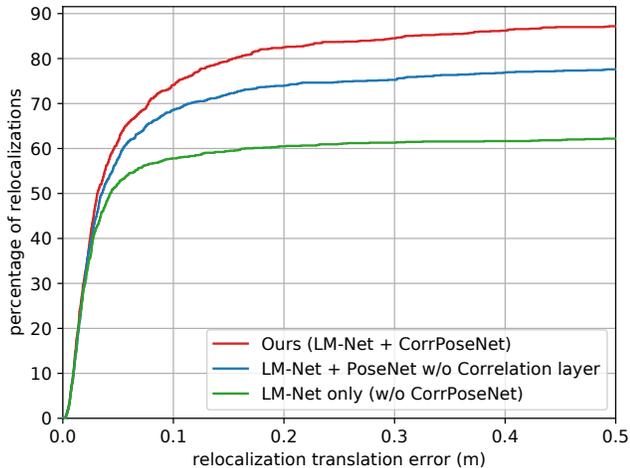
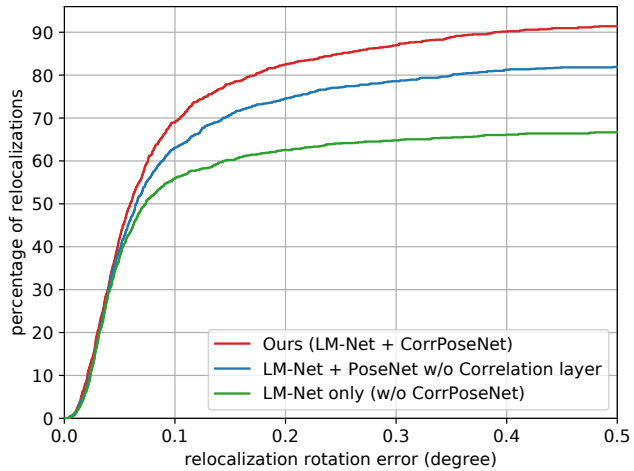

(a) Translation error.

(b) Rotation error.

**Figure 2:** Cumulative error plot for relocalization on the CARLA relocalization benchmark validation data [2]. It can be seen that the correlation layer in CorrPoseNet has a large impact on the performance.

**Table 2:** Network architecture and parameters of the regression block. **k** denotes kernel size, **s** stride, and **p** padding. $N_c = 256$ denotes the input channels for the CARLA model, and $N_o = 260$ denotes the input channels for the Oxford model, respectively.

| Regression block | | | | | | |
|---|---|---|---|---|---|---|
| layer | in-chns | out-chns | k | s | p | activation |
| conv0 | $N_c$ / $N_o$ | 128 | 7 | 1 | 0 | ReLU |
| BN | 128 | 128 | - | - | - | |
| conv1 | 128 | 64 | 5 | 1 | 0 | ReLU |
| BN | 64 | 64 | - | - | - | |
| FC | 2304 | 6 | - | - | - | |

**Table 3:** Output of the decoder of LM-Net. $H$, and $W$ denote height and width of the feature maps.

| Decoder layer | Output size |
|---|---|
| $F_1$ | $16 \times H/8 \times W/8$ |
| $F_2$ | $16 \times H/4 \times W/4$ |
| $F_3$ | $16 \times H/2 \times W/2$ |
| $F_4$ | $16 \times H \times W$ |

lutional block conv8 with 512 output channels, kernel size 3, stride 2, and padding 1. Otherwise the network architecture and parameters are the same as for CorrPoseNet. The results on the CARLA validation data are shown in Figure 2. Even the simpler pose estimation network (PoseNet w/o Correlation layer) improves the result over using identity as an initialization for the direct image alignment (LM-Net only). However, utilizing the correlation layer significantly boosts the performance.

**Table 4:** This table shows the AUC until 0.5 meters / 0.5 degrees for the relocalization error on the Oxford validation sequences. Our data augmentation (which warps the images using random poses) improves both rotation and translation error.

| Method | $t_{\text{AUC}}$ | $R_{\text{AUC}}$ |
|---|---|---|
| Ours | **80.45** | **65.11** |
| Ours w/o data augmentation | 80.15 | 64.58 |

## 4. Ablation Study Oxford Data Augmentation

We show the impact of the data augmentation for the Oxford RobotCar Relocalization benchmark, where we warp the images to different poses using dense depths in Table 4. It can be seen that the proposed augmentation improves translation and rotation error on the validation data.

## References

[1] O. Ronneberger, P. Fischer, and T. Brox. U-Net: Convolutional Networks for Biomedical Image Segmentation. In *MICCAI*, 2015. 1

[2] L. von Stumberg, P. Wenzel, Q. Khan, and D. Cremers. GN-Net: The Gauss-Newton Loss for Multi-Weather Relocalization. *RA-L*, 5(2):890–897, 2020. 2



\end{document}